\documentclass[sigconf]{acmart}

\usepackage{amsmath}
\usepackage{colortbl}
\usepackage{subcaption}
\usepackage{multirow}
\usepackage{algorithm}
\usepackage{algpseudocode}
\usepackage[capitalize]{cleveref}

\AtBeginDocument{%
  }

\setcopyright{none}
\renewcommand\footnotetextcopyrightpermission[1]{}
\acmConference[KDD '27]{ACM SIGKDD Conference on Knowledge Discovery and Data Mining}{August 1--5, 2027}{San Jose, CA, USA}
\acmYear{2027}
\copyrightyear{2027}

\newcommand{\method}{VDA}

\title{Visual Distribution Anchoring for Efficient Prompt Tuning}

\author{Pouya Parsa}
\affiliation{%
  \institution{University of Minnesota}
  \country{USA}
}

\author{Raoof Zare Moayedi}
\affiliation{%
  \institution{\mbox{}}
  \country{\mbox{}}
}

\author{Seongjin Choi}
\affiliation{%
  \institution{University of Minnesota}
  \country{USA}
}

\renewcommand{\shortauthors}{Parsa, Moayedi, and Choi}

\begin{document}

\begin{abstract}
  Prompt tuning adapts vision--language models with few trainable parameters, but
  existing designs face a trade-off: static textual prompts can overfit source
  classes, image-conditioned prompts incur per-instance computation, and
  multimodal tuning modifies the visual branch. We propose \method{} (Visual
  Distribution Anchoring), a training-free target adaptation framework that
  augments a frozen semantic classifier with class-level visual prototypes
  estimated offline from an unlabeled target pool. We first investigate whether
  such prototypes can be synthesized from class names alone. Although a
  text-to-centroid mapper accurately reconstructs held-out source prototypes, it
  does not transfer under dataset shift: class names specify semantic identity
  but not how classes are visually instantiated in a target domain. An oracle
  analysis nevertheless confirms that true target prototypes are highly
  discriminative. Motivated by this gap, \method{} uses frozen semantic and
  domain-template classifiers to hard-partition unlabeled target images into
  class-correlated groups. Confidence-ranked image features form normalized
  visual prototypes, which are fused with the semantic classifier using one
  global weight. Adaptation requires no target labels, target-side optimization,
  uniform class-prior assumption, iterative refinement, or access to test
  queries, and produces a fixed, cacheable classifier. Controlled experiments
  show that class-specific hard partitioning, rather than generic target-domain
  statistics, drives the improvement, and that visually local pseudo-label
  errors can remain useful despite being class-incorrect. Across ten
  ImageNet-to-target transfers, the same frozen design improves zero-shot CLIP,
  TCP, and MaPLe by \(3.22\), \(3.39\), and \(3.35\) points, respectively,
  improving nine of ten targets in every setting. Its visual correction further
  improves leakage-free PromptKD by \(2.79\) points, demonstrating
  complementarity across zero-shot, source-prompted, multimodal-prompted, and
  target-distilled classifiers.
  \end{abstract}


\maketitle

\section{Introduction}
  \label{sec:introduction}

  Vision--language models such as CLIP~\cite{radford2021learning} perform
  zero-shot recognition by comparing image representations with text classifiers
  constructed from class names. Prompt learning improves these classifiers using
  limited labeled source data
  \cite{zhou2022learning,zhou2022conditional,khattak2023maple,yao2024tcp},
  but introduces a trade-off between efficiency and adaptation. Static textual
  prompts are inexpensive but can overfit the source distribution;
  image-conditioned prompts adapt to individual queries at a per-instance cost;
  and multimodal methods modify both language and visual prompt paths. Despite
  these advances, a source-trained classifier may preserve a target class's
  semantic identity without capturing how that class is visually expressed in a
  new domain. For example, the name \emph{forest} identifies a concept but does
  not specify its appearance in an overhead satellite image.

  A natural solution is to predict a visual class prototype directly from each
  class name. We investigate this possibility by mapping frozen class-name
  embeddings to image-space centroids. A regularized linear mapper accurately
  reconstructs centroids for held-out source classes, indicating that text
  embeddings contain substantial information about general class appearance.
  However, the resulting prototypes do not improve cross-dataset recognition.
  Their raw similarity to target centroids is dominated by shared feature
  structure, while their mean-centered alignment with the target class geometry
  remains weak. Thus, accurately reconstructing source-domain prototypes does
  not imply recovering how the same semantic classes appear in a different
  domain.

  This failure does not mean that visual prototypes are uninformative. An oracle
  analysis shows the opposite: across five development datasets, direct
  classification with true 16-shot target centroids increases mean accuracy from
  \(49.91\%\) to \(66.16\%\). Conservative fusion of the visual and semantic
  classifiers also produces a large gain, whereas routing the same true
  centroids through a class-aware prompt generator captures only a small
  fraction of their signal. These findings reveal two distinct limitations:
  class names alone do not determine domain-specific target prototypes, and
  indirect prompt conditioning may not effectively exploit visual prototypes
  even when they are available.

  We propose \method{} (Visual Distribution Anchoring),\linebreak a training-free target
  adaptation framework that augments a frozen semantic classifier with
  class-level visual prototypes estimated offline from an unlabeled target pool.
  The target-training pool is disjoint from the evaluation set, and its labels
  are not used. A frozen semantic classifier and a fixed domain-template
  classifier first score every unlabeled image. Their logits are averaged, and
  hard argmax assignment partitions the pool into class-correlated visual groups.
  Within each group, VDA ranks images by assignment confidence and retains at
  most \(K\) supports. Their normalized mean forms a target visual prototype,
  which is fused with the corresponding semantic classifier using a single
  global weight. Classes receiving no assignments retain their original
  semantic classifier.

  The VDA procedure operates on classifier and prototype logits rather than the
  internal architecture of a particular prompt learner. It requires no target
  labels, target-side gradient optimization, validation-based stopping, uniform
  class-prior assumption, iterative pseudo-label refinement, or access to test
  queries. When the semantic and visual logits share a feature space, their
  fusion reduces to one fixed class vector that can be cached before evaluation.
  VDA therefore adds target visual information without altering source prompt
  training or generating new prompts for each query.

  Hard partitioning is central to the method. Balanced soft assignment forces
  the target pool toward a prescribed class distribution and can distribute
  nearly identical evidence across multiple prototypes. In our experiments, it
  produces high-entropy assignments and progressively worse prototypes under
  iterative refinement. VDA instead assigns each image to one predicted class
  and allows unequal class masses. The support budget \(K\) is only a cap:
  classes with many assignments retain a confidence-ranked subset, classes with
  fewer assignments retain all available evidence, and unsupported classes use
  the semantic classifier alone.

  \begin{figure*}[t]
      \centering
      \includegraphics[width=0.93\linewidth]{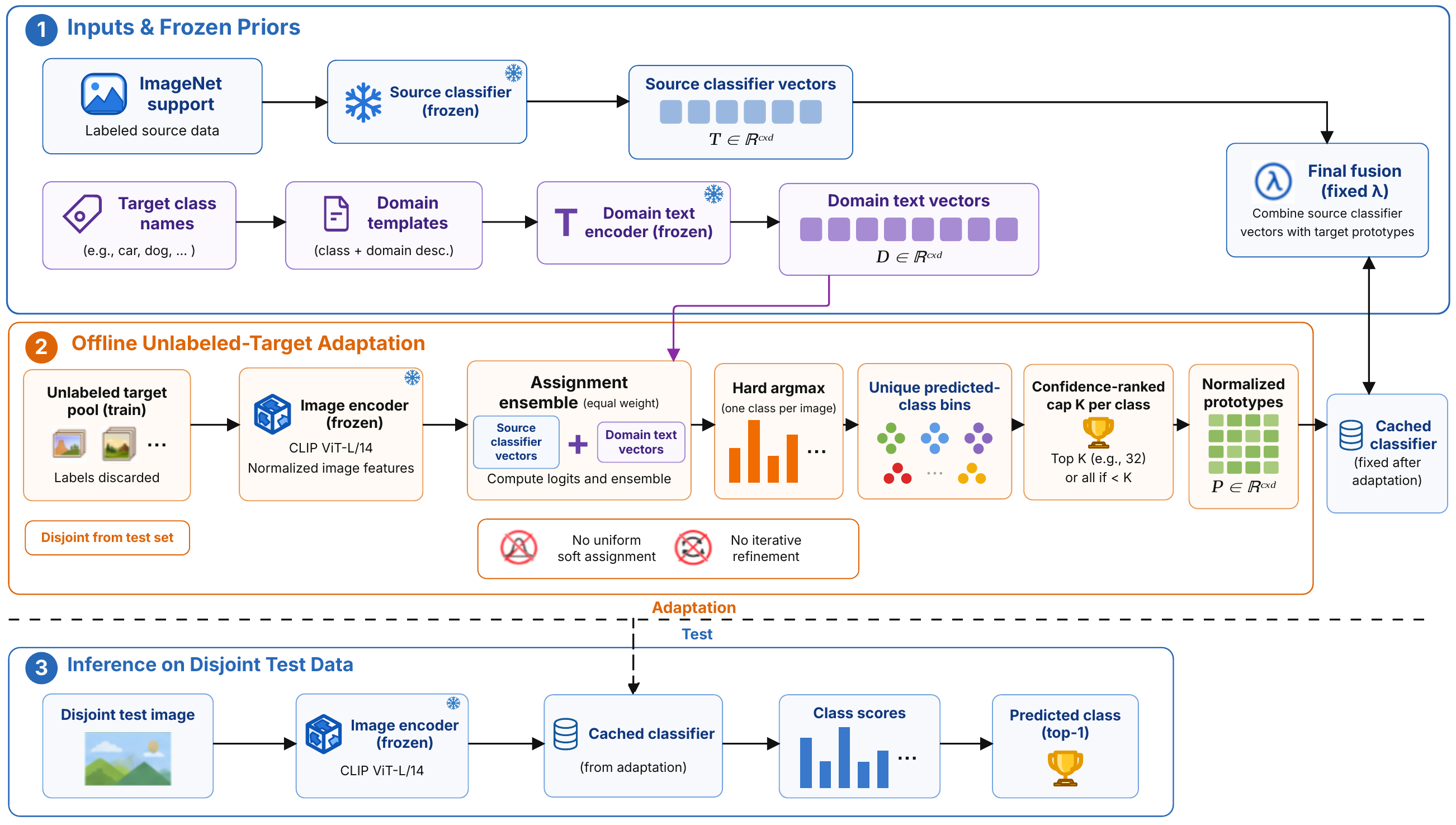}
      \caption{
          \textbf{Overview of \method{}.}
          Frozen semantic and domain-template classifiers hard-partition a
          disjoint unlabeled target pool. Confidence-ranked supports form
          normalized class-specific visual prototypes, which are fused with the
          semantic classifiers before evaluation. VDA uses no target labels,
          target-side optimization, uniform class-prior assumption, iterative
          refinement, or test queries.
      }
      \label{fig:overview}
  \end{figure*}

  Importantly, useful target evidence does not require perfectly accurate
  pseudo-labels. Our controls show that hard class correspondence accounts for
  most of the improvement, while confidence ordering provides a smaller and
  domain-dependent benefit. Incorrect assignments can remain useful when the
  predicted and true classes occupy nearby visual neighborhoods. On
  FGVC-Aircraft, for example, selected errors are substantially more similar to
  the true class than random wrong-class pairs and are strongly enriched for the
  correct aircraft family and manufacturer. Such images provide relevant
  fine-grained and target-domain appearance even when they do not preserve the
  exact class identity. Semantic--visual fusion exploits this imperfect evidence
  conservatively: the prototype supplies target appearance, while the semantic
  classifier retains class specificity.

  Across ten ImageNet-to-target transfers, VDA improves a strong source-trained
  prompt classifier from \(65.82\%\) to \(69.21\%\), improving nine of ten
  targets. Under a matched protocol using the same unlabeled target pool and
  disjoint test split, it exceeds sparse inductive ZLaP
  \cite{stojnic2024label} by \(1.53\) points and prior-free InMaP
  \cite{qian2024intramodal} by \(2.98\) points. It obtains a similar mean to
  uniform-prior InMaP without target-proxy optimization or a specified target
  class distribution.

  We further test whether the effect is tied to one source prompt architecture.
  Using the same \(K\), fusion weight, domain templates, and one-pass adaptation
  design, a fully standalone zero-shot CLIP variant improves from \(65.34\%\) to
  \(68.56\%\). The same fixed visual correction improves multimodal MaPLe
  \cite{khattak2023maple} from \(66.25\%\) to \(69.60\%\) and leakage-free
  PromptKD~\cite{li2024promptkd} from \(70.80\%\) to \(73.59\%\). Each setting
  improves nine of ten targets. These results show that class-level target visual
  information remains complementary across zero-shot, source-prompted,
  multimodal-prompted, and target-distilled semantic classifiers.

  Our contributions are threefold:
  \begin{itemize}
      \item We distinguish source prototype reconstruction from target visual
      transfer. Name-to-centroid mapping accurately recovers held-out source
      geometry but does not predict domain-specific target prototypes, while
      oracle experiments establish that true target prototypes contain
      substantial discriminative information.

      \item We introduce \method{}, a training-free target adaptation framework
      that estimates class-correlated visual prototypes from a disjoint
      unlabeled target pool and fuses them with a frozen semantic classifier.
      VDA requires no target labels, target optimization, uniform class-prior
      assumption, iterative refinement, or test-time adaptation.

      \item Across ten datasets and multiple semantic classifiers, we show that
      class-specific hard partitioning---rather than generic target-domain
      statistics---drives the improvement, and that visually local pseudo-label
      errors can remain useful when combined with a semantically stable
      classifier.
  \end{itemize}

\section{Related Work}
  \label{sec:related_work}

  \paragraph{Prompt adaptation for vision--language models.}
  CLIP performs zero-shot recognition using text classifiers constructed from
  class names and hand-crafted templates \cite{radford2021learning}. CoOp
  replaces manually selected context words with continuous prompt tokens
  \cite{zhou2022learning},\linebreak while CoCoOp generates image-conditioned prompts to
  improve transfer at the cost of per-instance computation
  \cite{zhou2022conditional}. KgCoOp and PromptSRC preserve knowledge from the
  pretrained model to reduce source overfitting
  \cite{yao2023visual,khattak2023selfregulating}; MaPLe and MMRL instead adapt
  both language and visual prompt paths
  \cite{khattak2023maple,guo2025mmrl}. TCP, our primary source classifier,
  generates class-aware prompts from textual class knowledge
  \cite{yao2024tcp}. Prompt design is also important in specialized
  applications. Medical and remote-sensing methods use pathology-specific or
  image-conditioned prompts
  \cite{tiu2022expert,koleilat2025biomedcoop,singha2023applenet}, while recent
  vehicle-surveillance work evaluates task-specific and self-reflective prompts
  for make and model recognition
  \cite{parsa2025videobasedvehiclesurveillancewild}. Such prompting can encode
  valuable domain knowledge but must often be designed and validated for each
  task. VDA is complementary: it estimates class-level visual evidence from an
  unlabeled deployment pool and fuses it with an existing semantic classifier,
  without changing its prompt architecture or generating prompts per query.

  \paragraph{Semantic-to-visual transfer.}
  Predicting visual representations from semantic descriptions has been studied
  in zero-shot learning. Visual-exemplar methods map seen-class semantics to
  visual prototypes for unseen categories \cite{changpinyo2017predicting}, while
  feature-synthesis approaches generate visual examples from class semantics
  \cite{long2017zeroshot}. SuS-X similarly constructs synthetic or retrieved
  support sets from target category names to adapt frozen vision--language
  models \cite{udandarao2023susx}. We investigate whether CLIP class names can
  directly predict image-space centroids, but find that accurate source-centroid
  reconstruction does not recover domain-specific target geometry. VDA therefore
  estimates prototypes from real unlabeled target images. The prototypes provide
  target appearance, while the frozen semantic classifier retains class identity
  when pseudo-support assignments are imperfect.

  \paragraph{Unlabeled-target adaptation.}
  Test-Time Prompt Tuning adapts prompts by minimizing entropy over augmented
  query views \cite{shu2022testtime}, and PromptAlign additionally aligns source
  and target feature statistics \cite{samadh2023align}. These approaches perform
  optimization during inference. PromptKD instead distills a larger teacher into
  a target-specific student using unlabeled target images
  \cite{li2024promptkd}. More closely related to VDA, InMaP initializes and
  optimizes visual proxies from unlabeled target features and can impose a
  uniform target prior through balanced assignment
  \cite{qian2024intramodal}. ZLaP propagates labels over a text--image graph and
  retains graph-based scoring for new queries \cite{stojnic2024label}. VDA
  differs by performing a single training-free adaptation pass: frozen
  classifiers hard-partition the unlabeled pool, capped class-specific prototypes
  are estimated without assuming a uniform prior, and semantic--visual fusion
  produces a fixed classifier before evaluation. It requires no target gradient
  updates, query-time graph propagation, iterative pseudo-label refinement, or
  access to test queries.

\section{Visual Distribution Anchoring}
  \label{sec:method}

  \subsection{Problem Setting}
  \label{sec:problem_setting}

  Let \(\mathcal{C}=\{1,\ldots,C\}\) denote a target class vocabulary and
  \(\mathcal{U}=\{x_i\}_{i=1}^{N}\) an unlabeled target-training pool. The pool
  is disjoint from the evaluation set. We seek to adapt a frozen semantic
  classifier using \(\mathcal{U}\), without target labels, target-side gradient
  updates, validation-based stopping, or access to test queries during
  adaptation.

  VDA uses a frozen CLIP image encoder \(f_V\) to represent the target visual
  distribution. Each target image is encoded and unit-normalized as
  \begin{equation}
      v_i
      =
      \operatorname{norm}\!\left(f_V(x_i)\right)
      =
      \frac{f_V(x_i)}{\lVert f_V(x_i)\rVert_2}
      \in\mathbb{R}^{d}.
      \label{eq:normalized_image}
  \end{equation}
  Let \(s^{\mathrm{sem}}(x,c)\) denote the score assigned to image \(x\) and
  class \(c\) by a frozen semantic classifier. This classifier may be a
  zero-shot text classifier or a source-trained prompt model. In our primary
  experiments, it is instantiated with TCP~\cite{yao2024tcp}, but VDA does not
  use TCP's internal prompt generator.

  A class name constrains semantic identity but does not uniquely determine how
  the class appears in a target domain. VDA therefore estimates target visual
  prototypes from \(\mathcal{U}\) and uses them to provide a class-specific
  appearance correction to \(s^{\mathrm{sem}}\).

  \subsection{Domain-Aware Hard Assignment}
  \label{sec:hard_assignment}

  A bare class name may not fully describe the target image domain. We construct
  a second frozen classifier from a fixed domain template. Let \(\pi_D(c)\)
  combine class \(c\) with a known domain description, such as ``a close-up
  photograph of a \(c\) texture'' or ``a true-color overhead satellite image of
  \(c\).'' Its normalized text vector is
  \begin{equation}
      d_c
      =
      \operatorname{norm}\!\left(f_T(\pi_D(c))\right),
      \label{eq:domain_text_vector}
  \end{equation}
  where \(f_T\) is the frozen CLIP text encoder. The corresponding domain score
  is
  \begin{equation}
      s^{D}(x_i,c)
      =
      \gamma\,v_i^\top d_c,
      \label{eq:domain_score}
  \end{equation}
  with frozen CLIP logit scale \(\gamma\). The domain description is fixed before
  target evaluation and does not depend on target labels. Exact templates are
  provided in the supplementary material.

  For every unlabeled image, VDA averages the semantic and domain-template
  scores:
  \begin{equation}
      a_{ic}
      =
      \frac{1}{2}
      \left[
          s^{\mathrm{sem}}(x_i,c)
          +
          s^{D}(x_i,c)
      \right].
      \label{eq:assignment_score}
  \end{equation}
  The image receives one hard assignment,
  \begin{equation}
      \hat c_i
      =
      \arg\max_{c\in\mathcal{C}} a_{ic}.
      \label{eq:hard_argmax}
  \end{equation}
  This produces disjoint predicted-class partitions
  \begin{equation}
      \mathcal{I}_c
      =
      \{i:\hat c_i=c\}.
      \label{eq:class_partition}
  \end{equation}

  Hard assignment has two useful properties. First, every target image can
  contribute to at most one visual prototype. Second, the partition sizes are
  allowed to be unequal. VDA therefore does not force a uniform target-class
  prior or distribute every image fractionally across multiple classes.

  \subsection{Capped Pseudo-Support Selection}
  \label{sec:pseudo_support}

  Within each predicted class, images are ranked by assignment confidence. We
  define
  \begin{equation}
      q_i
      =
      \operatorname{softmax}(a_i)_{\hat c_i}
      =
      \frac{\exp(a_{i\hat c_i})}
           {\sum_{j\in\mathcal{C}}\exp(a_{ij})}.
      \label{eq:assignment_confidence}
  \end{equation}
  For class \(c\), VDA retains
  \begin{equation}
      \mathcal{S}_c
      =
      \operatorname{TopK}_{i\in\mathcal{I}_c}(q_i),
      \qquad
      |\mathcal{S}_c|
      =
      \min(K,|\mathcal{I}_c|).
      \label{eq:topk_selection}
  \end{equation}

  The support budget \(K\) is a cap rather than a balancing constraint. If fewer
  than \(K\) images are assigned to a class, all assigned images are retained.
  If \(\mathcal{I}_c=\varnothing\), the class receives no visual prototype and
  later falls back to its semantic classifier. We use \(K=32\), selected on the
  development datasets and frozen for all subsequent evaluations.

  The selected images are pseudo-supports rather than assumed ground-truth
  examples. Exact pseudo-label correctness is not required: an incorrectly
  assigned image can remain useful if its visual representation lies near the
  target class or a closely related fine-grained category.

  \subsection{Target Visual Prototypes}
  \label{sec:visual_prototypes}

  For every supported class, we estimate a normalized target visual prototype:
  \begin{equation}
      \hat{\mu}_c
      =
      \operatorname{norm}
      \left(
          \frac{1}{|\mathcal{S}_c|}
          \sum_{i\in\mathcal{S}_c}v_i
      \right).
      \label{eq:visual_prototype}
  \end{equation}
  The prototype represents the first moment of the class-correlated target
  features selected by the frozen semantic and domain priors. Individual image
  features are normalized before averaging, and the resulting mean is normalized
  again.

  VDA performs this estimation once. The resulting prototypes are not used to
  iteratively relabel the target pool. Avoiding assignment--prototype
  iteration prevents weak initial assignments from being repeatedly reinforced.

  \subsection{Semantic--Visual Fusion}
  \label{sec:semantic_visual_fusion}

  Pseudo-support prototypes capture target appearance but may not preserve exact
  semantic identity. VDA therefore treats them as corrections to the semantic
  classifier rather than standalone replacements.

  Define a class-dependent fusion coefficient
  \begin{equation}
      \lambda_c
      =
      \begin{cases}
          \lambda, & |\mathcal{S}_c|>0,\\
          0,       & |\mathcal{S}_c|=0,
      \end{cases}
      \label{eq:class_fusion_weight}
  \end{equation}
  where \(\lambda=0.2\) is selected once on the development datasets and fixed
  across all targets. For a query image \(x\), the visual-prototype score is
  \begin{equation}
      s^{\mathrm{vis}}(x,c)
      =
      \gamma\,
      \operatorname{norm}\!\left(f_V(x)\right)^\top
      \hat{\mu}_c.
      \label{eq:visual_score}
  \end{equation}
  The final VDA score is
  \begin{equation}
      s^{\mathrm{VDA}}(x,c)
      =
      (1-\lambda_c)s^{\mathrm{sem}}(x,c)
      +
      \lambda_c s^{\mathrm{vis}}(x,c).
      \label{eq:vda_score}
  \end{equation}
  Prediction is performed by
  \begin{equation}
      \hat y(x)
      =
      \arg\max_{c\in\mathcal{C}}
      s^{\mathrm{VDA}}(x,c).
      \label{eq:vda_prediction}
  \end{equation}

  We deliberately use one global fusion weight.\linebreak Class-dependent reliability
  weights can place different classes on incompatible score scales and make the
  multiclass argmax poorly calibrated. A fixed global weight instead provides a
  conservative correction across classes and datasets.

  \subsection{Shared-Space and Cross-Classifier Forms}
  \label{sec:classifier_forms}

  \paragraph{Shared CLIP feature space.}
  When the semantic classifier uses the same normalized CLIP image feature as
  the visual prototypes, its score can be written as
  \begin{equation}
      s^{\mathrm{sem}}(x,c)
      =
      \gamma\,v(x)^\top w_c^{\mathrm{sem}},
      \label{eq:shared_semantic_score}
  \end{equation}
  where \(w_c^{\mathrm{sem}}\) is its class vector. This includes our primary TCP
  and standalone zero-shot CLIP settings. Equation~\eqref{eq:vda_score} then
  reduces to
  \begin{equation}
      s^{\mathrm{VDA}}(x,c)
      =
      \gamma\,v(x)^\top\bar w_c,
      \label{eq:single_linear_score}
  \end{equation}
  with
  \begin{equation}
      \bar w_c
      =
      \begin{cases}
          (1-\lambda)w_c^{\mathrm{sem}}
          +\lambda\hat{\mu}_c,
          & |\mathcal{S}_c|>0,\\[2mm]
          w_c^{\mathrm{sem}},
          & |\mathcal{S}_c|=0.
      \end{cases}
      \label{eq:fused_class_vector}
  \end{equation}
  The fused vector is not renormalized because
  Eq.~\eqref{eq:fused_class_vector} implements logit interpolation. The
  \(\{\bar w_c\}\) vectors are cached before evaluation, and inference requires
  one standard image encoding and one classifier matrix multiplication.

  \paragraph{Cross-classifier attachment.}
  VDA can also provide a fixed visual correction to a semantic classifier whose
  query logits are produced in another adapted feature space. Let
  \(\tilde{s}^{\mathrm{sem}}(x,c)\) denote logits from such a classifier, such as
  MaPLe or PromptKD. We retain the target prototypes estimated in the frozen
  CLIP space and apply
  \begin{equation}
      \tilde{s}^{\mathrm{VDA}}(x,c)
      =
      (1-\lambda_c)\tilde{s}^{\mathrm{sem}}(x,c)
      +
      \lambda_c s^{\mathrm{vis}}(x,c).
      \label{eq:cross_classifier_attachment}
  \end{equation}
  This form tests whether the target visual correction is complementary to a
  different semantic learner. It remains training-free with respect to VDA, but
  may require both the attached model's logits and the frozen CLIP visual logits
  at inference. We therefore distinguish these attachment experiments from the
  single-feature-space form in Eq.~\eqref{eq:single_linear_score}.

  \subsection{Adaptation Procedure and Complexity}
  \label{sec:adaptation_complexity}

  VDA consists of one offline pass over the unlabeled target pool:

  \begin{enumerate}
      \item encode and normalize all target-training images;
      \item compute semantic and domain-template assignment scores;
      \item hard-partition the pool by argmax;
      \item retain at most \(K\) confidence-ranked features per class;
      \item average and normalize the selected features;
      \item fuse the resulting visual scores with the semantic classifier.
  \end{enumerate}

  After image features are extracted, assignment requires
  \(O(NCd)\) similarity computation. Top-\(K\) selection can be implemented
  with per-class size-\(K\) heaps, requiring \(O(N\log K)\) additional work.
  A streaming implementation retains at most \(O(CKd)\) selected feature values,
  and the final prototype matrix requires \(O(Cd)\) memory.

  VDA introduces no trainable target parameters, optimizer, validation stopping
  criterion, target-prior constraint, or assignment-refinement loop. In the
  shared-space setting, the final classifier has the same asymptotic inference
  cost and storage as an ordinary frozen CLIP classifier.

\section{Experiments}
  \label{sec:experiments}

  \subsection{Experimental Setup}
  \label{sec:experimental_setup}

  \paragraph{Datasets.}
  We train source prompt models on ImageNet and evaluate cross-dataset transfer
  on Caltech101, Oxford Pets, Stanford Cars, Flowers102, Food101,
  FGVC-Aircraft, SUN397, DTD, EuroSAT, and UCF101. For each dataset, the
  complete target-training split forms an unlabeled adaptation pool, and accuracy
  is measured on the disjoint target-test split. Target-training labels are not
  accessed during adaptation. They are used only in explicitly marked post-hoc
  analyses of pseudo-label precision and prototype geometry.

  DTD, EuroSAT, UCF101, Aircraft, and Cars were used for design selection,
  including the support budget \(K\) and fusion weight \(\lambda\). Preliminary
  results from an earlier \(K=16\) configuration had also been viewed on the
  remaining datasets. We therefore report all ten datasets uniformly as
  cross-dataset evaluations and do not characterize any subset as a pristine
  untouched test set.

  \paragraph{Backbone and semantic classifiers.}
  All methods use OpenAI CLIP ViT-B/16 unless otherwise noted. Our primary
  semantic classifier is TCP~\cite{yao2024tcp}, trained independently on the
  corresponding 16-shot ImageNet splits for seeds 1--3. The resulting ImageNet
  top-1 accuracies are \(71.66\%\), \(71.50\%\), and \(71.50\%\). To test
  whether VDA depends on one prompt architecture, we additionally apply it to
  seven-template zero-shot CLIP, three official ImageNet-trained
  MaPLe~\cite{khattak2023maple} checkpoints, and three leakage-free
  PromptKD~\cite{li2024promptkd} checkpoints.

  \paragraph{VDA configuration.}
  VDA averages logits from a frozen semantic classifier and a fixed
  domain-template classifier, assigns every unlabeled image by hard argmax, and
  ranks images by maximum-softmax confidence within each predicted class. It
  retains at most \(K=32\) images per class, averages their unit-normalized CLIP
  features, and renormalizes the resulting visual prototype. Prototype logits
  are fused with semantic logits using one global weight \(\lambda=0.2\).
  Classes receiving no assignments retain their semantic classifier. We use the
  complete unlabeled pool and perform no iterative assignment or prototype
  refinement. The domain templates, \(K\), \(\lambda\), normalization,
  aggregation, and selection rule are fixed across datasets.

  \paragraph{Baseline selection and protocol matching.}
  Table~\ref{tab:main_cross_dataset} is organized by information access rather
  than as an unrestricted leaderboard. Zero-shot CLIP and TCP are the unadapted
  semantic references. ZLaP~\cite{stojnic2024label} and
  InMaP~\cite{qian2024intramodal} are included because they can use the same
  unlabeled target pool and represent two complementary adaptation strategies:
  graph-based label propagation and visual-proxy optimization.

  Methods using different resources are evaluated separately. MaPLe is a
  source-trained multimodal prompt method, while PromptKD uses a larger ViT-L/14
  teacher and target-specific student distillation. They are evaluated in the
  classifier-generality analysis in Sec.~\ref{sec:classifier_generality}.\linebreak
  Published MMRL and MMRL+ReBaPL results
  \cite{guo2025mmrl,bendou2026rebapl} are reported in the supplementary material
  as source-only context. This protocol-stratified presentation avoids treating
  methods with different target-data, optimization, and model budgets as
  directly equivalent.

  \paragraph{InMaP protocol.}
  The primary downstream experiments in InMaP estimate visual proxies using
  access to the target test image collection and apply dataset-specific
  pseudo-label refinement. Our protocol instead estimates target prototypes from
  the unlabeled target-training split and reserves the test split for evaluation.
  We therefore do not transcribe InMaP's published \(70.46\%\) ten-dataset mean
  into Table~\ref{tab:main_cross_dataset} as though it were a matched result.

  We report a matched prior-free reproduction in which InMaP optimizes its visual
  proxies on the same unlabeled target-training pool used by VDA. A separate
  uniform-marginal Sinkhorn sensitivity is reported in the supplementary
  material, but it is not labeled as the official InMaP configuration: the
  published method estimates and smooths a reference class distribution and uses
  dataset-specific refinement parameters. The published per-dataset results are
  also retained in the supplementary material with an explicit
  protocol-different marker. A matched reproduction using the complete official
  refinement procedure will replace the prior-free row if it can be validated
  under the disjoint-pool protocol.

  \paragraph{PromptKD protocol.}
  The released PromptKD cross-dataset implementation assigns the target test
  split to its validation field while selecting the best validation checkpoint.
  To prevent target-test checkpoint selection, we preserve the authors'
  optimizer, learning rate, augmentations, teacher, and fixed 20-epoch schedule,
  but evaluate only the final checkpoint. We run all ten datasets using three
  independently trained last-step checkpoints. The published \(71.33\%\)
  cross-dataset mean is retained only as protocol-different context; our
  classifier-generality analysis uses the leakage-free last-step results.

  \subsection{Main Cross-Dataset Comparison}
  \label{sec:main_cross_dataset}

  Table~\ref{tab:main_cross_dataset} presents the main matched comparison. VDA
  improves its primary semantic classifier from \(65.82\%\) to \(69.21\%\), a
  gain of \(3.39\) points, and improves nine of ten datasets. Food101 is
  essentially unchanged, decreasing by \(0.12\) points.

  VDA exceeds ZLaP by \(1.53\) points on average and obtains higher accuracy on
  eight datasets. It exceeds the matched prior-free InMaP reproduction by
  \(2.98\) points and obtains higher accuracy on nine datasets. These comparisons
  show that hard prototype adaptation can outperform graph propagation and
  optimized visual proxies without requiring an assumed target class
  distribution.

  The published InMaP result is numerically stronger than our prior-free
  reproduction, reporting a \(70.46\%\) mean over the same ten dataset names.
  However, because it estimates proxies under a different target-image protocol
  and uses dataset-specific refinement, it is not ranked against VDA in
  Table~\ref{tab:main_cross_dataset}. We report it separately rather than
  conflating accuracy differences with differences in target-data access and
  adaptation procedure.

  \begin{table*}[t]
      \centering
      \caption{
          Cross-dataset top-1 accuracy (\%) with OpenAI CLIP ViT-B/16.
          Zero-shot CLIP and TCP are unadapted semantic references. ZLaP, the
          matched InMaP reproduction, and VDA receive the same unlabeled
          target-training pool and are evaluated on the disjoint test split
          without target-training labels. Published InMaP results use a different
          target-image and refinement protocol and are reported separately.
          Best matched result in each column is bold.
      }
      \label{tab:main_cross_dataset}
      \setlength{\tabcolsep}{3.0pt}
      \scriptsize
      \resizebox{0.92\textwidth}{!}{
      \begin{tabular}{lccccccccccc}
          \toprule
          Method
          & Caltech & Pets & Cars & Flowers & Food
          & Aircraft & SUN & DTD & Euro. & UCF
          & Mean \\
          \midrule

          Zero-shot CLIP
          & 93.91 & 88.31 & 66.15 & 66.95 & 85.70
          & 23.04 & 66.25 & 45.15 & 50.36 & 67.54
          & 65.34 \\

          TCP~\cite{yao2024tcp}
          & 94.20 & 89.79 & 64.79 & 71.32 & 86.36
          & 24.29 & 67.10 & 44.46 & 48.18 & 67.72
          & 65.82 \\
          \midrule

          ZLaP~\cite{stojnic2024label}
          & 92.05 & 86.26 & 66.06 & 72.59 & \textbf{87.33}
          & \textbf{25.89} & 67.22 & 48.82 & 58.22 & 72.30
          & 67.68 \\

          InMaP, matched, no refinement~\cite{qian2024intramodal}
          & 94.28 & 87.08 & 66.80 & 70.85 & 86.77
          & 22.83 & 67.31 & 45.51 & 53.21 & 67.70
          & 66.23 \\
          \midrule

          \method{}
          & \textbf{95.05} & \textbf{90.78} & \textbf{67.91}
          & \textbf{75.25} & 86.24 & 25.26
          & \textbf{70.18} & \textbf{48.98}
          & \textbf{59.84} & \textbf{72.59}
          & \textbf{69.21} \\
          \bottomrule
      \end{tabular}}
  \end{table*}

  \paragraph{Visual prototypes require semantic fusion.}
  The direct hard-\(K\) prototype classifier obtains a ten-dataset mean of
  \(65.92\%\), only \(0.10\) points above the primary semantic classifier.\linebreak
  Semantic--visual fusion reaches \(69.21\%\). Target prototypes therefore
  contain useful appearance information but do not reliably preserve exact
  class identity when used alone. VDA treats each visual prototype as a
  correction: the semantic classifier retains class identity, while the
  prototype contributes target-specific appearance.

  \subsection{Generality Across Semantic Classifiers}
  \label{sec:classifier_generality}

  VDA operates on semantic and visual-prototype logits rather than the internal\linebreak
  prompt architecture. We evaluate this property across zero-shot,
  source-prompted, multimodal-prompted, and target-distilled classifiers.
  Table~\ref{tab:classifier_generality} summarizes the results.

  \begin{table}[t]
      \centering
      \caption{
          VDA generality across semantic classifiers. The zero-shot row uses the
          fully standalone VDA procedure. MaPLe and PromptKD use the same fixed
          seed-1 TCP/domain assignments and raw-CLIP visual correction across
          their three model seeds; these rows test correction compatibility
          rather than matched training or inference cost.
      }
      \label{tab:classifier_generality}
      \setlength{\tabcolsep}{4.3pt}
      \small
      \resizebox{\columnwidth}{!}{%
      \begin{tabular}{lcccc}
          \toprule
          Semantic classifier
          & Base & \(+\) VDA & Gain & Improved \\
          \midrule

          Zero-shot CLIP
          & 65.34 & 68.56 & \(+3.22\) & 9/10 \\

          TCP
          & 65.82 & 69.21 & \(+3.39\) & 9/10 \\

          MaPLe, 3 seeds
          & \(66.25{\pm}.42\)
          & \(69.60{\pm}.22\)
          & \(+3.35{\pm}.37\)
          & 9/10 \\

          PromptKD, 3 seeds
          & \(70.80{\pm}.40\)
          & \(73.59{\pm}.28\)
          & \(+2.79{\pm}.38\)
          & 9/10 \\
          \bottomrule
      \end{tabular}}
  \end{table}

  {\sloppy
  \paragraph{Standalone zero-shot CLIP.}\mbox{}\\
  For the strict classifier-independence test, zero-shot CLIP replaces TCP in\linebreak
  assignment, confidence ranking, zero-assignment fallback, and final fusion.
  The domain templates, equal-logit assignment rule, \(K=32\), \(\lambda=0.2\),
  and all other VDA choices remain unchanged. Accuracy increases from
  \(65.34\%\) to \(68.56\%\), a \(3.22\)-point gain, and nine of ten datasets
  improve. Food101 changes by only \(-0.05\) points. This establishes that VDA
  does not depend on TCP's internal prompt generator.
  }

  As an attachment sensitivity, we also retain the primary TCP/domain
  assignments and fuse their visual correction with zero-shot CLIP. This
  condition reaches \(68.61\%\), nearly identical to the standalone result.
  Because TCP remains in its assignment path, its per-dataset values are
  reported in the supplementary material rather than used as the primary
  generality result.

  \paragraph{Multimodal prompt transfer.}
  We attach the same fixed seed-1 visual correction to three official
  ImageNet-trained MaPLe checkpoints. MaPLe modifies both language and visual
  prompt paths, providing a stronger architectural transfer test than a static
  zero-shot classifier. Its mean increases from \(66.25\pm0.42\%\) to
  \(69.60\pm0.22\%\), a paired gain of \(3.35\pm0.37\) points. Nine datasets
  improve, while Food101 is unchanged (\(-0.01\pm0.16\)).

  \paragraph{Target-distilled classifier.}
  We similarly combine the fixed raw-CLIP prototype logits with three
  leakage-free PromptKD checkpoints. PromptKD increases from
  \(70.80\pm0.40\%\) to \(73.59\pm0.28\%\), a paired gain of
  \(2.79\pm0.38\) points. Every seed improves nine datasets, while Food101
  decreases by \(0.51\pm0.12\) points.

  The MaPLe and PromptKD experiments are logit-level attachments: semantic
  logits from the attached model are combined with visual-prototype logits
  estimated in the raw CLIP space. They demonstrate that the VDA visual
  correction remains complementary across prompt architectures and adaptation
  regimes, but they do not imply matched optimization or inference budgets.
  PromptKD, in particular, uses a larger teacher and target-specific student
  training.

  Together, these results show that class-level target visual information is
  useful across the tested semantic classifiers: zero-shot CLIP, source textual
  prompt tuning, multimodal prompt tuning, and target-specific prompt
  distillation all benefit from VDA.

  \subsection{Diagnosing Visual Prototype Transfer}
  \label{sec:prototype_diagnostics}

  \paragraph{Name-only prototype prediction.}
  We first test whether target prototypes can be synthesized without target
  images. An affine ridge mapper is trained from ImageNet class-name embeddings
  to normalized 16-shot source image centroids. Across three class-disjoint
  \(700/150/150\) splits, it obtains held-out centroid cosine between \(0.922\)
  and \(0.933\), substantially exceeding an MLP mapper. Five-fold out-of-fold
  predictions achieve \(0.930\) source-centroid cosine, yet their downstream
  mean is \(49.49\%\), below the \(49.91\%\) semantic baseline.

  The discrepancy is explained by target-domain geometry. Predicted prototypes
  have raw cosine \(0.814\) to true target centroids but only \(0.387\)
  mean-centered cosine, \(17.8\%\) retrieval top-1 accuracy, and pairwise
  Spearman correlation \(0.488\). Raw similarity therefore overstates recovery
  of class-specific target structure. Accurate source-domain reconstruction is
  not sufficient for cross-domain prototype transfer.

  \paragraph{Oracle target prototypes.}
  True target prototypes are nevertheless strongly discriminative. With
  disjoint 16-shot target support, direct centroid classification reaches
  \(66.16\%\), compared with \(49.91\%\) for the semantic classifier.
  Fixed \(\lambda=0.2\) fusion reaches \(60.52\%\). In contrast, providing the
  same true centroids to the original class-aware prompt-generator interface
  reaches only \(50.48\%\). Five-fold target cross-fitting produces the same
  qualitative result while excluding every query from its centroid. These
  experiments motivate estimating target prototypes from unlabeled target images
  and combining them directly with semantic logits.

  \paragraph{Failure of balanced soft assignment.}
  We also evaluate class-balanced Sinkhorn assignment over the complete
  unlabeled pool. At round 0, direct prototypes obtain \(36.52\%\), while
  fusion reaches \(50.09\%\). After three rounds of reassignment, they decrease
  to \(27.91\%\) and \(49.90\%\), respectively. Normalized assignment entropy
  remains between \(0.949\) and \(0.994\), indicating that each prototype
  receives nearly the same mixture of target images. Hard partitioning avoids
  this collapse by assigning each image to one predicted class, retaining only a
  capped subset, and making no uniform-prior assumption.

  \subsection{Ablation and Mechanism Analysis}
  \label{sec:ablation}

  \paragraph{Source-seed repeatability.}\mbox{}\\
  Across three independently trained primary semantic classifiers, VDA obtains
  \(54.92\pm0.28\%\) on the five datasets used for design selection, compared
  with \(49.89\pm0.23\%\) before adaptation. Every source-seed--dataset pair
  improves. The seed-level mean gains are approximately \(+4.70\), \(+4.93\),
  and \(+5.46\) points. Aircraft is the least certain dataset, but its mean
  change remains positive.

  \paragraph{Support budget.}
  For \(K\in\{1,2,4,8,16,32\}\), fusion obtains \(49.86\), \(51.35\),
  \(53.07\), \(54.21\), \(54.80\), and \(54.92\%\), respectively. Direct
  prototype accuracy continues to improve with \(K\), but semantic--visual
  fusion saturates between \(K=16\) and \(K=32\): the selected \(K=32\)
  configuration exceeds \(K=16\) by only \(0.12\) points.

  \paragraph{Class-specificity controls.}
  Table~\ref{tab:class_specificity} isolates the source of the visual correction.
  A common target centroid preserves the semantic classifier's argmax because it
  adds the same score to every class. Random target images and shuffled
  prototypes fail, showing that generic target-domain information does not
  explain the gain. Random sampling within each hard-predicted class reaches
  \(53.23\%\), while confidence ordering reaches \(54.92\%\). Hard class
  correspondence therefore supplies most of the improvement, with confidence
  ranking adding a smaller \(1.69\)-point aggregate benefit.

  \begin{table}[t]
      \centering
      \caption{
          Class-specificity controls on the five datasets used for design
          selection. Random conditions use five repetitions; deviations are
          over repetition-level aggregate means.
      }
      \label{tab:class_specificity}
      \setlength{\tabcolsep}{4.2pt}
      \small
      \begin{tabular}{lcc}
          \toprule
          Prototype construction & Direct & Fusion \\
          \midrule

          Common target centroid
          & 3.19 & 49.89 \\

          Shuffled prototypes
          & \(2.15{\pm}.97\) & \(37.78{\pm}1.03\) \\

          Random target assignment
          & \(2.94{\pm}.85\) & \(48.57{\pm}.54\) \\

          Random within predicted class
          & \(51.35{\pm}.24\) & \(53.23{\pm}.24\) \\

          Confidence-ranked within class
          & \textbf{52.15} & \textbf{54.92} \\
          \bottomrule
      \end{tabular}
  \end{table}

  The confidence-ranking effect is heterogeneous: it contributes \(+5.80\)
  points on EuroSAT and \(+1.70\) on UCF101 but is statistically
  indistinguishable from zero on Aircraft. We therefore attribute the consistent
  mechanism to hard class partitioning rather than confidence ranking alone.

  \paragraph{Domain-template attribution.}
  The fixed domain-template classifier obtains \(50.30\%\), compared with
  \(49.89\%\) for the primary semantic classifier. Using semantic-only,
  domain-only, and equal semantic/domain rankings for visual adaptation yields
  \(54.23\%\), \(54.82\%\), and \(54.92\%\), respectively. Domain-aware text
  improves target assignment, but the larger increase from the domain-template
  classifier to domain-ranked visual fusion shows that selected target image
  features provide the dominant correction.

  \paragraph{Visually useful pseudo-label errors.}
  Exact pseudo-label precision is lowest on Aircraft (\(32.9\%\)), yet VDA
  still improves its classifier. Incorrect selected supports have raw
  true-centroid cosine \(0.960\), compared with \(0.861\pm0.081\) for random
  wrong-class pairs. After mean centering, the comparison is \(0.653\) versus
  \(0.003\). Selected errors lie at the \(88.7\)th percentile of wrong-pair
  visual similarity on average. They share the true aircraft family \(17.1\%\)
  of the time, compared with \(1.25\%\) by chance, and share the manufacturer
  \(38.4\%\) of the time, compared with \(7.98\%\) by chance. Thus, exact
  pseudo-label accuracy understates prototype utility: visually local errors can
  provide relevant target appearance while the semantic classifier preserves
  exact class identity.

  \paragraph{Unlabeled-pool size and class coverage.}
  Using nested label-blind subsets of \(10\%\), \(25\%\), \(50\%\), and
  \(100\%\) of the target pool yields means of \(51.03\%\), \(53.16\%\),
  \(54.15\%\), and \(54.92\%\), compared with \(49.89\%\) before adaptation.
  Fifty percent is the smallest audited fraction that improves all five
  design-selection datasets, while the complete pool performs best on every
  dataset.

  The nominal \(K=32\) is an upper bound rather than an attained count. All
  EuroSAT classes reach 32 supports, but only approximately \(45\%\) of
  Aircraft and \(53\%\) of Cars classes do so. Four to five Aircraft classes and
  one to two Cars classes receive no assignments. Such classes retain their
  semantic classifier, although their predictions can still change when
  competing classes receive visual corrections. VDA therefore primarily refines
  classes already recognized by the semantic initializer and cannot reliably
  recover classes receiving no target support.

\section{Discussion}
  \label{sec:discussion}

  \paragraph{Semantic and visual information are complementary.}
  Accurate source-domain text-to-centroid reconstruction does not yield
  transferable target prototypes because class names specify semantic identity,
  not how classes appear under a particular domain shift. Nevertheless, oracle
  target centroids are highly discriminative. VDA bridges this gap by estimating
  visual prototypes from unlabeled target images and treating them as corrections
  rather than replacements for the semantic classifier. The semantic component
  preserves class identity, while the visual prototype adjusts the classifier
  toward target-domain appearance.

  \paragraph{Class correspondence matters more than exact pseudo-labels.}
  A common target centroid, random target assignment, and shuffled prototypes do
  not reproduce VDA's gain, showing that generic target-domain statistics are
  insufficient. Most of the improvement comes from hard class partitioning;
  confidence ranking provides a smaller and domain-dependent benefit. Exact
  pseudo-label accuracy is also an incomplete measure of prototype quality. On
  Aircraft, for example, many incorrect supports remain visually close to the
  true class and are enriched for the correct family or manufacturer. Such
  examples can contribute useful fine-grained appearance even when they do not
  identify the exact variant.

  \paragraph{Generality across semantic classifiers.}
  VDA improves zero-shot CLIP, TCP, MaPLe, and leakage-free PromptKD by
  \(3.22\), \(3.39\), \(3.35\), and \(2.79\) points, respectively. These gains
  indicate that class-level target visual evidence remains complementary across
  zero-shot, source-prompted, multimodal-prompted, and target-distilled
  classifiers. Under the matched disjoint-pool protocol, VDA also outperforms
  ZLaP and our prior-free InMaP reproduction without target optimization,
  query-time graph propagation, or an assumed target class prior. Published
  InMaP results use a different target-image and refinement protocol and are
  therefore not treated as a matched comparison.

  \paragraph{Efficiency and deployment.}
  VDA performs one frozen adaptation pass and requires no parameter updates or
  iterative refinement. When semantic and visual logits share the CLIP feature
  space, their fusion reduces to one cached class vector with ordinary linear
  inference. Cross-feature-space attachments such as MaPLe and PromptKD still
  benefit from the fixed visual correction, although they may require an
  additional raw-CLIP score stream. VDA is therefore most attractive when an
  unlabeled deployment pool is available but target labels, repeated
  optimization, and per-query adaptation are impractical.

  \section{Limitations}
  \label{sec:limitations}

  \paragraph{Dependence on unlabeled target images.}
  VDA is not a strict source-only or name-only method. It assumes access to an
  unlabeled target-training pool representative of the evaluation domain.
  Performance may degrade when this pool is small, biased, temporally mismatched,
  or missing some target classes. Although VDA improves the average with partial
  pools, full-pool access performs best in our experiments.

  \paragraph{Dependence on a domain description.}
  Assignment combines the semantic classifier with a manually specified
  domain-template classifier. The description is coarse and requires no target
  labels, but remains an additional form of prior knowledge. When the deployment
  domain is unknown, heterogeneous, or difficult to describe, selecting an
  appropriate template may require manual effort.

  \paragraph{Raw CLIP visual space.}
  All target prototypes are estimated using OpenAI CLIP ViT-B/16. While we
  evaluate zero-shot CLIP, TCP, MaPLe, and PromptKD semantic classifiers, these
  experiments do not establish transfer to arbitrary backbones. MaPLe and
  PromptKD are also logit-level attachments whose semantic outputs are combined
  with visual logits computed in the raw CLIP space; they may therefore require
  an additional CLIP score stream at inference.

  \paragraph{Coverage of difficult classes.}
  The support budget \(K=32\) is an upper bound rather than an attained count for
  every class. Fine-grained datasets contain classes with few or no predicted
  assignments, and VDA cannot reliably recover classes receiving no relevant
  pseudo-support. Retaining the semantic classifier for a zero-support class also
  does not guarantee unchanged accuracy because corrections applied to competing
  classes can alter the multiclass decision boundary.

  \paragraph{Fixed global fusion.}
  A single fusion weight is stable and avoids cross-class score inconsistency,
  but cannot fully exploit unusually accurate prototypes. Our class-dependent
  reliability experiments degraded performance, so we retain a global weight and
  leave reliable label-free fusion estimation as future work.

  \paragraph{Evaluation and baseline protocols.}
  Design choices were selected using five target datasets, and preliminary
  results from an earlier support budget had been viewed on the remaining
  datasets. The results should therefore be interpreted as broad cross-dataset
  evidence rather than a fully untouched benchmark. Published methods also do
  not always share the same target-data protocol. In particular, published
  InMaP results use a different target-image and refinement setting, while
  PromptKD uses a larger teacher and target-specific optimization. We separate
  such results from matched comparisons, but these differences prevent a fully
  resource-equivalent ranking.

  \section{Conclusion}
  \label{sec:conclusion}

  {\emergencystretch=1em
  \looseness=-1
  We presented \method{} (Visual Distribution Anchoring), a training-free target
  adaptation framework that augments a frozen semantic classifier with
  class-level visual prototypes estimated from an unlabeled target pool. We first
  showed that accurate source-domain text-to-centroid reconstruction does not
  produce transferable target prototypes, even though oracle target centroids
  are strongly discriminative. VDA addresses this gap by hard-partitioning
  unlabeled target images, estimating capped class-specific prototypes, and
  combining their visual logits with a semantically stable classifier.\par}

  {\emergencystretch=1em
  \looseness=-1
  Across ten ImageNet-to-target datasets, VDA improves its primary semantic
  classifier from \(65.82\%\) to \(69.21\%\) and outperforms matched ZLaP and
  prior-free InMaP reproductions. The same frozen adaptation design improves
  standalone zero-shot CLIP by \(3.22\) points, while its visual correction
  improves MaPLe and leakage-free PromptKD by \(3.35\) and \(2.79\) points,
  respectively. Controlled analyses show that class-specific hard assignment,
  rather than generic target-domain statistics, drives the improvement, and that
  visually local pseudo-label errors can remain useful despite being
  class-incorrect.\par}

  {\emergencystretch=1em
  \looseness=-1
  VDA requires no target labels, target-side optimization, uniform class-prior
  assumption, iterative refinement, or access to evaluation queries during
  adaptation. When the semantic and visual classifiers share a feature space,
  the result is a fixed, cacheable classifier with ordinary linear inference. We
  hope these findings encourage further study of class-level visual evidence as
  a simple and complementary mechanism for adapting prompt-based
  vision--language models to new domains.\par}

\bibliographystyle{ACM-Reference-Format}
\bibliography{refs}

@inproceedings{radford2021learning,
  title         = {Learning Transferable Visual Models From Natural Language Supervision},
  author        = {Radford, Alec and Kim, Jong Wook and Hallacy, Chris and Ramesh, Aditya and Goh, Gabriel and Agarwal, Sandhini and Sastry, Girish and Askell, Amanda and Mishkin, Pamela and Clark, Jack and Krueger, Gretchen and Sutskever, Ilya},
  booktitle     = {Proceedings of the 38th International Conference on Machine Learning},
  series        = {Proceedings of Machine Learning Research},
  volume        = {139},
  pages         = {8748--8763},
  publisher     = {PMLR},
  year          = {2021},
  url           = {https://proceedings.mlr.press/v139/radford21a.html}
}

@article{zhou2022learning,
  title         = {Learning to Prompt for Vision-Language Models},
  author        = {Zhou, Kaiyang and Yang, Jingkang and Loy, Chen Change and Liu, Ziwei},
  journal       = {International Journal of Computer Vision},
  volume        = {130},
  number        = {9},
  pages         = {2337--2348},
  year          = {2022},
  doi           = {10.1007/s11263-022-01653-1},
  url           = {https://doi.org/10.1007/s11263-022-01653-1}
}

@inproceedings{zhou2022conditional,
  title         = {Conditional Prompt Learning for Vision-Language Models},
  author        = {Zhou, Kaiyang and Yang, Jingkang and Loy, Chen Change and Liu, Ziwei},
  booktitle     = {Proceedings of the IEEE/CVF Conference on Computer Vision and Pattern Recognition (CVPR)},
  pages         = {16816--16825},
  year          = {2022},
  doi           = {10.1109/CVPR52688.2022.01631},
  url           = {https://openaccess.thecvf.com/content/CVPR2022/html/Zhou_Conditional_Prompt_Learning_for_Vision-Language_Models_CVPR_2022_paper.html}
}

@inproceedings{stojnic2024label,
  title         = {Label Propagation for Zero-Shot Classification with Vision-Language Models},
  author        = {Stojni{\'c}, Vladan and Kalantidis, Yannis and Tolias, Giorgos},
  booktitle     = {Proceedings of the IEEE/CVF Conference on Computer Vision and Pattern Recognition},
  pages         = {23209--23218},
  year          = {2024},
  url           = {https://openaccess.thecvf.com/content/CVPR2024/html/Stojni_Label_Propagation_for_Zero-shot_Classification_with_Vision-Language_Models_CVPR_2024_paper.html}
}

@inproceedings{qian2024intramodal,
  title         = {Intra-Modal Proxy Learning for Zero-Shot Visual Categorization with {CLIP}},
  author        = {Qian, Qi and Xu, Yuanhong and Hu, Juhua},
  booktitle     = {Advances in Neural Information Processing Systems},
  volume        = {36},
  pages         = {25461--25474},
  year          = {2023},
  url           = {https://proceedings.neurips.cc/paper_files/paper/2023/hash/50a057e9fe79ffa3f4120fb6fb88071a-Abstract-Conference.html}
}

@inproceedings{khattak2023maple,
  title         = {MaPLe: Multi-modal Prompt Learning},
  author        = {Khattak, Muhammad Uzair and Rasheed, Hanoona and Maaz, Muhammad and Khan, Salman and Khan, Fahad Shahbaz},
  booktitle     = {Proceedings of the IEEE/CVF Conference on Computer Vision and Pattern Recognition (CVPR)},
  pages         = {19113--19122},
  year          = {2023},
  doi           = {10.1109/CVPR52729.2023.01832},
  url           = {https://openaccess.thecvf.com/content/CVPR2023/html/Khattak_MaPLe_Multi-Modal_Prompt_Learning_CVPR_2023_paper.html}
}

@inproceedings{khattak2023selfregulating,
  title         = {Self-regulating Prompts: Foundational Model Adaptation without Forgetting},
  author        = {Khattak, Muhammad Uzair and Wasim, Syed Talal and Naseer, Muzammal and Khan, Salman and Yang, Ming-Hsuan and Khan, Fahad Shahbaz},
  booktitle     = {Proceedings of the IEEE/CVF International Conference on Computer Vision (ICCV)},
  pages         = {15190--15200},
  year          = {2023},
  doi           = {10.1109/ICCV51070.2023.01394},
  url           = {https://openaccess.thecvf.com/content/ICCV2023/html/Khattak_Self-regulating_Prompts_Foundational_Model_Adaptation_without_Forgetting_ICCV_2023_paper.html}
}

@inproceedings{yao2023visual,
  title         = {Visual-Language Prompt Tuning with Knowledge-guided Context Optimization},
  author        = {Yao, Hantao and Zhang, Rui and Xu, Changsheng},
  booktitle     = {Proceedings of the IEEE/CVF Conference on Computer Vision and Pattern Recognition (CVPR)},
  pages         = {6757--6767},
  year          = {2023},
  url           = {https://openaccess.thecvf.com/content/CVPR2023/html/Yao_Visual-Language_Prompt_Tuning_With_Knowledge-Guided_Context_Optimization_CVPR_2023_paper.html}
}

@inproceedings{yao2024tcp,
  title         = {TCP: Textual-based Class-aware Prompt Tuning for Vision-Language Model},
  author        = {Yao, Hantao and Zhang, Rui and Xu, Changsheng},
  booktitle     = {Proceedings of the IEEE/CVF Conference on Computer Vision and Pattern Recognition (CVPR)},
  pages         = {23438--23448},
  year          = {2024},
  doi           = {10.1109/CVPR52733.2024.02212},
  url           = {https://openaccess.thecvf.com/content/CVPR2024/html/Yao_TCP_Textual-Based_Class-Aware_Prompt_Tuning_for_Visual-Language_Model_CVPR_2024_paper.html}
}

@inproceedings{guo2025mmrl,
  title         = {MMRL: Multi-Modal Representation Learning for Vision-Language Models},
  author        = {Guo, Yuncheng and Gu, Xiaodong},
  booktitle     = {Proceedings of the IEEE/CVF Conference on Computer Vision and Pattern Recognition (CVPR)},
  pages         = {25015--25025},
  year          = {2025},
  url           = {https://openaccess.thecvf.com/content/CVPR2025/html/Guo_MMRL_Multi-Modal_Representation_Learning_for_Vision-Language_Models_CVPR_2025_paper.html}
}

@inproceedings{li2024promptkd,
  title         = {PromptKD: Unsupervised Prompt Distillation for Vision-Language Models},
  author        = {Li, Zheng and Li, Xiang and Fu, Xinyi and Zhang, Xin and Wang, Weiqiang and Chen, Shuo and Yang, Jian},
  booktitle     = {Proceedings of the IEEE/CVF Conference on Computer Vision and Pattern Recognition (CVPR)},
  pages         = {26617--26626},
  year          = {2024},
  url           = {https://openaccess.thecvf.com/content/CVPR2024/html/Li_PromptKD_Unsupervised_Prompt_Distillation_for_Vision-Language_Models_CVPR_2024_paper.html}
}

@inproceedings{bendou2026rebapl,
  title         = {ReBaPL: Repulsive Bayesian Prompt Learning},
  author        = {Bendou, Yassir and Ezzahir, Omar and Montesuma, Eduardo and Mahuas, Gabriel and Schevchenko, Victoria and Gartrell, Mike},
  booktitle     = {Proceedings of the IEEE/CVF Conference on Computer Vision and Pattern Recognition (CVPR)},
  year          = {2026},
  url           = {https://openaccess.thecvf.com/content/CVPR2026/html/Bendou_ReBaPL_Repulsive_Bayesian_Prompt_Learning_CVPR_2026_paper.html}
}

@inproceedings{changpinyo2017predicting,
  title     = {Predicting Visual Exemplars of Unseen Classes for Zero-Shot Learning},
  author    = {Changpinyo, Soravit and Chao, Wei-Lun and Sha, Fei},
  booktitle = {Proceedings of the IEEE International Conference on Computer Vision (ICCV)},
  pages     = {3476--3485},
  year      = {2017},
  url       = {https://openaccess.thecvf.com/content_iccv_2017/html/Changpinyo_Predicting_Visual_Exemplars_ICCV_2017_paper.html}
}

@inproceedings{long2017zeroshot,
  title     = {From Zero-Shot Learning to Conventional Supervised Classification: Unseen Visual Data Synthesis},
  author    = {Long, Yang and Liu, Li and Shao, Ling and Shen, Fumin and Ding, Guiguang and Han, Jungong},
  booktitle = {Proceedings of the IEEE Conference on Computer Vision and Pattern Recognition (CVPR)},
  pages     = {1627--1636},
  year      = {2017},
  url       = {https://openaccess.thecvf.com/content_cvpr_2017/html/Long_From_Zero-Shot_Learning_CVPR_2017_paper.html}
}

@inproceedings{udandarao2023susx,
  title     = {{SuS-X}: Training-Free Name-Only Transfer of Vision-Language Models},
  author    = {Udandarao, Vishaal and Gupta, Ankush and Albanie, Samuel},
  booktitle = {Proceedings of the IEEE/CVF International Conference on Computer Vision (ICCV)},
  pages     = {2725--2736},
  year      = {2023},
  url       = {https://openaccess.thecvf.com/content/ICCV2023/html/Udandarao_SuS-X_Training-Free_Name-Only_Transfer_of_Vision-Language_Models_ICCV_2023_paper.html}
}

@inproceedings{shu2022testtime,
  title     = {Test-Time Prompt Tuning for Zero-Shot Generalization in Vision-Language Models},
  author    = {Shu, Manli and Nie, Weili and Huang, De-An and Yu, Zhiding and Goldstein, Tom and Anandkumar, Anima and Xiao, Chaowei},
  booktitle = {Advances in Neural Information Processing Systems},
  volume    = {35},
  pages     = {14274--14289},
  year      = {2022},
  doi       = {10.52202/068431-1038},
  url       = {https://proceedings.neurips.cc/paper_files/paper/2022/hash/5bf2b802e24106064dc547ae9283bb0c-Abstract-Conference.html}
}

@inproceedings{samadh2023align,
  title     = {Align Your Prompts: Test-Time Prompting with Distribution Alignment for Zero-Shot Generalization},
  author    = {Abdul Samadh, Jameel and Gani, Mohammad Hanan and Hussein, Noor and Khattak, Muhammad Uzair and Naseer, Muhammad Muzammal and Khan, Fahad Shahbaz and Khan, Salman H.},
  booktitle = {Advances in Neural Information Processing Systems},
  volume    = {36},
  year      = {2023},
  url       = {https://proceedings.neurips.cc/paper_files/paper/2023/hash/fe8debfd5a36ada52e038c8b2078b2ce-Abstract-Conference.html}
}

@inproceedings{parsa2025videobasedvehiclesurveillancewild,
  title     = {Video-Based Vehicle Surveillance in the Wild: License Plate, Make, and Model Recognition with Self-Reflective Vision-Language Models},
  author    = {Parsa, Pouya and Li, Keya and Kockelman, Kara M. and Choi, Seongjin},
  booktitle = {Proceedings of the 105th Annual Meeting of the Transportation Research Board},
  address   = {Washington, DC},
  year      = {2026},
  note      = {Paper TRBAM-26-03946},
  url       = {https://www.caee.utexas.edu/prof/kockelman/public_html/TRB26VideoSurveillance.pdf}
}

@article{tiu2022expert,
  title     = {Expert-Level Detection of Pathologies from Unannotated Chest X-Ray Images via Self-Supervised Learning},
  author    = {Tiu, Ekin and Talius, Ellie and Patel, Pujan and Langlotz, Curtis P. and Ng, Andrew Y. and Rajpurkar, Pranav},
  journal   = {Nature Biomedical Engineering},
  volume    = {6},
  pages     = {1399--1406},
  year      = {2022},
  publisher = {Springer Nature},
  doi       = {10.1038/s41551-022-00936-9},
  url       = {https://www.nature.com/articles/s41551-022-00936-9}
}

@inproceedings{koleilat2025biomedcoop,
  title     = {BiomedCoOp: Learning to Prompt for Biomedical Vision-Language Models},
  author    = {Koleilat, Taha and Asgariandehkordi, Hojat and Rivaz, Hassan and Xiao, Yiming},
  booktitle = {Proceedings of the IEEE/CVF Conference on Computer Vision and Pattern Recognition (CVPR)},
  pages     = {14766--14776},
  year      = {2025},
  url       = {https://openaccess.thecvf.com/content/CVPR2025/html/Koleilat_BiomedCoOp_Learning_to_Prompt_for_Biomedical_Vision-Language_Models_CVPR_2025_paper.html}
}

@inproceedings{singha2023applenet,
  title     = {{APPLeNet}: Visual Attention Parameterized Prompt Learning for Few-Shot Remote Sensing Image Generalization Using {CLIP}},
  author    = {Singha, Mainak and Jha, Ankit and Solanki, Bhupendra and Bose, Shirsha and Banerjee, Biplab},
  booktitle = {Proceedings of the IEEE/CVF Conference on Computer Vision and Pattern Recognition (CVPR) Workshops},
  pages     = {2024--2034},
  year      = {2023},
  url       = {https://openaccess.thecvf.com/content/CVPR2023W/EarthVision/html/Jha_APPLeNet_Visual_Attention_Parameterized_Prompt_Learning_for_Few-Shot_Remote_Sensing_CVPRW_2023_paper.html}
}

\end{document}